\documentclass[10pt,a4paper]{article}
\usepackage[scale=.80]{geometry}
\usepackage[affil-it]{authblk}
\usepackage{amssymb,amsmath,amsthm,amsbsy}
\usepackage{physics} 
\usepackage[pdftex,bookmarksnumbered]{hyperref}
\usepackage{graphicx}
\usepackage{epstopdf}
\usepackage{epsfig}
\usepackage{multicol}
\usepackage{subfigure}
\usepackage{multirow}
\usepackage{algorithm,algorithmic}

\usepackage{setspace}
\usepackage{color}
\usepackage{lineno}

\usepackage{tcolorbox}
\usepackage[width=.8\textwidth]{caption}
\usepackage{lscape}

\usepackage{times}
\title{\Large\textbf{A brief note on understanding neural networks as Gaussian processes}}
\author{Mengwu Guo\thanks{E-mail: \texttt{m.guo@utwente.nl}}}
\affil{Applied Analysis, Department of Applied Mathematics, University of Twente}
\date{}

\begin{document}

\maketitle 
\thispagestyle{empty}

\noindent\textit{Abstract}: As a generalization of the work in \cite{lee2017deep} (Lee et al., 2017), this note briefly discusses when the prior of a neural network output follows a Gaussian process, and how a neural-network-induced Gaussian process is formulated. The posterior mean functions of such a Gaussian process regression lie in the reproducing kernel Hilbert space defined by the neural-network-induced kernel. In the case of two-layer neural networks, the induced Gaussian processes provide an interpretation of the reproducing kernel Hilbert spaces whose union forms a Barron space.

\vspace{1mm}


\section{Neural network}

We consider an $L$-hidden-layer fully-connected neural network \cite{strang2019linear} with width $N_l$ and nonlinear activation function $\phi$ for the $l$-th layer, $1\leq l \leq L$. At the $j$-th neuron in the $l$-th layer of the network, the pre-bias and post-activation are denoted by $z_j^{[l]}$ and $x_j^{[l]}$, respectively, $1\leq i \leq N_l$. Let $\vb*{x}=\vb*{x}^{[0]}\in\mathbb{R}^{d_\text{in}}$ denote the inputs of the network and $\vb*{y}=\vb*{z}^{[L+1]}\in\mathbb{R}^{d_\text{out}}$ denote the outputs. Note that we also let $N_0=d_\text{in}$ and $N_{L+1}=d_\text{out}$. Weight and bias parameters between the $(l-1)$-th and $l$-th layers are represented by $W_{ij}^{[l]}$ and $b_i^{[l]}$, respectively, $1\leq l\leq (L+1)$, $1\leq i \leq N_{l}$, $1\leq j \leq N_{l-1}$. Then one has
\begin{equation}\label{def}
\begin{split}
& z_i^{[l]}(\vb*{x})=\sum_{j=1}^{N_{l-1}}W_{ij}^{[l]} x_j^{[l-1]}(\vb*{x})\,,\quad
x_i^{[l]}(\vb*{x})=\phi(z_i^{[l]}(\vb*{x})+b_i^{[l]} )\,,\quad 1\leq i \leq N_l, ~1\leq l \leq L\,, \quad \text{and}\\
& y_i(\vb*{x})=z_i^{[L+1]}(\vb*{x})= \sum_{j=1}^{N_{L}}W_{ij}^{[L+1]} x_j^{[L]}(\vb*{x})\,,\quad  1\leq i \leq d_\text{out}\,.
\end{split}
\end{equation}

A multivariate function $\vb*{f}: \Omega \to \mathbb{R}^{d_\text{out}}$ ($\Omega\subset \mathbb{R}^{d_\text{in}}$) is approximated by a vector-valued surrogate of neural network, denoted by $\vb*{f}^\text{NN}(\cdot;\vb*{W},\vb*{b}):\Omega\to \mathbb{R}^{d_\text{out}}$. Here,  vectors $\vb*{W}$ and $\vb*{b}$ collect all the weight and bias parameters, respectively. Based on the data of $M$ input-output pairs $\{(\vb*{x}^{(m)},\vb*{y}^{(m)})\}_{m=1}^{M}$, the training of such a neural network is often performed by minimizing a loss function:
\begin{equation}
(\vb*{W},\vb*{b})=\arg\min_{\vb*{W},\vb*{b}} \left\{ \frac{1}{M}\sum_{m=1}^{M}\|\vb*{y}^{(m)}-\vb*{f}^\text{NN}(\vb*{x}^{(m)};\vb*{W},\vb*{b})\|_2^2 + \lambda \|\vb*{W}\|_2^2\right\}\,,
\end{equation}
in which the first term is the \textit{mean square error} and the second is an $L_2$-regularization term with $\lambda \geq 0$ being the penalty coefficient.

\section{Neural network prior as Gaussian processes}\label{NNGP}

The prior of a neural network output is a Gaussian process under the following assumptions \cite{lee2017deep,neal2012bayesian}\footnote{The discussion in this section generalizes the work in \cite{lee2017deep}.}:

\begin{itemize}
	\item For the first layer ($l=1$), $(\vb*{w}_i^{[1]},b_i^{[1]})\sim \pi$ are independent and identically distributed (i.i.d) with respect to $1\leq i \leq N_1$, $\vb*{w}_i^{[1]}:=\left\{W_{i1}^{[1]},W_{i2}^{[1]},\cdots,W_{id_\text{in}}^{[1]}\right\}^\text{T}$, and $\pi$ can be any distribution;
	
	\item All the other weights and biases are independently drawn;
	
	\item For the $l$-th layer, $2\leq l \leq L+1$, $W_{ij}^{[l]}$'s are i.i.d. with mean value $\mu^{[l]}_w/N_{l-1}$ and variance ${\sigma_w^2}^{[l]} / N_{l-1}$, $1\leq i \leq N_{l}$, $1\leq j \leq N_{l-1}$;
	
	\item For the $l$-th layer, $2\leq l \leq L$, $b_{i}^{[l]}$'s are i.i.d. with mean value $\mu^{[l]}_b$ and variance ${\sigma_b^2}^{[l]} $, $1\leq i \leq N_{l}$ ; $b_i^{[L+1]}=0$, $1\leq i \leq d_\text{out}$; and

	\item $N_l \to \infty$, $1\leq l \leq L$.
\end{itemize}
First we show by induction that $\{z^{[l]}_i(\vb*{x})|~i\in \mathbb{N}^+\}$ are independent, identical random fields over $\Omega$, $1\leq l \leq L+1$. When $l=1$, $z_i^{[1]}(\vb*{x})={\vb*{w}_i^{[1]}}^\text{T}\vb*{x}$, and the proposition apparently holds true. Assume it holds true for $l$, $1\leq l \leq L$. We consider two arbitrary locations $\vb*{x}$, $\vb*{x}'\in \Omega$ and hence have
\begin{equation}
\mathbb{E}[z_i^{[l+1]}(\vb*{x})]= \mu_w^{[l+1]} \mathbb{E}_{z_\cdot^{[l]},b_\cdot^{[l]}}[\phi(z_\cdot^{[l]}(\vb*{x})+b_\cdot^{[l]})]\,,
\end{equation}
and
\begin{equation}
\begin{split}
& ~\text{Cov}[z_i^{[l+1]}(\vb*{x}), z_j^{[l+1]}(\vb*{x}')]  \\
= & \sum_{p,q}\text{Cov}[W_{ip}^{[l+1]}x_p^{[l]}(\vb*{x}), W_{jq}^{[l+1]}x_q^{[l]}(\vb*{x}')]\\
 = & \sum_{p,q} \left\{\mathbb{E}[W_{ip}^{[l+1]}W_{jq}^{[l+1]}] \cdot \mathbb{E}[x_p^{[l]}(\vb*{x})x_q^{[l]}(\vb*{x}')]-\mathbb{E}[W_{ip}^{[l+1]}]\cdot\mathbb{E}[W_{jq}^{[l+1]}] \cdot \mathbb{E}[x_p^{[l]}(\vb*{x})]\cdot\mathbb{E}[x_q^{[l]}(\vb*{x}')]\right\}\\
 = & \sum_{p} \left\{\mathbb{E}[W_{ip}^{[l+1]}W_{jp}^{[l+1]}] \cdot \mathbb{E}[x_p^{[l]}(\vb*{x})x_p^{[l]}(\vb*{x}')]-\frac{{\left(\mu_w^{[l+1]}\right)}^2}{N_l^2}\cdot \mathbb{E}[x_p^{[l]}(\vb*{x})]\cdot\mathbb{E}[x_p^{[l]}(\vb*{x}')]\right\}\\
& - \sum_{p\neq q} \frac{{\left(\mu_w^{[l+1]}\right)}^2}{N_l^2} \cdot \underbrace{\text{Cov}[x_p^{[l]}(\vb*{x}),x_q^{[l]}(\vb*{x}')]}_{=0,\text{ since }z_p^{[l]}\text{ and }z_q^{[l]}\text{ are independent when }p\neq q} \\
= & ~ \frac{{\left(\mu_w^{[l+1]}\right)}^2}{N_l}~\text{Cov}[x_\cdot^{[l]}(\vb*{x}),x_\cdot^{[l]}(\vb*{x}')] + {\sigma_w^2}^{[l+1]}\delta_{ij} \mathbb{E}[x_\cdot^{[l]}(\vb*{x}),x_\cdot^{[l]}(\vb*{x}')]\\
\longrightarrow & ~ {\sigma_w^2}^{[l+1]}\delta_{ij} \mathbb{E}_{z_\cdot^{[l]},b_\cdot^{[l]}}[\phi(z_\cdot^{[l]}(\vb*{x})+b_\cdot^{[l]})\cdot\phi(z_\cdot^{[l]}(\vb*{x}')+b_\cdot^{[l]})]\,,\quad \text{as  } N_l \to \infty\,.
\end{split}
\end{equation}
Here $\delta_{ij}$ denotes the Kronecker delta. When $i\neq j$, the covariance value equals zero, implying that $\{z^{[l+1]}_i(\vb*{x})|~i\in \mathbb{N}^+\}$ are all uncorrelated. In fact, they are independent as each $z_\cdot^{[l+1]}$ is independently defined as a combination of $\{x_i^{[l]}|i\in\mathbb{N}^+\}$. It is also trivial to see that these random fields are identical, meaning that the proposition holds true for $l+1$ as well, which completes its proof.

As defined in \eqref{def}, furthermore, $z_\cdot^{[l]}$ equals a sum of $N_{l-1}$ i.i.d. random variables whose mean and variance are proportional to $1/N_{l-1}$. As $N_{l-1}\to \infty$, the central limit theorem \cite{durrett2019probability} can be applied, which gives that $\{z^{[l]}_i(\vb*{x})|~i\in \mathbb{N}^+\}$ follow identical, independent Gaussian processes, $2\leq l \leq L+1$, i.e., $z_\cdot^{[l]}\sim \mathcal{GP}(h^{[l]},k^{[l]})$ whose mean and covariance functions are
\begin{equation}\label{recursive}
\begin{split}
h^{[l]}(\vb*{x}) & =\mu_w^{[l]} \mathbb{E}_{z_\cdot^{[l-1]},b_\cdot^{[l-1]}}[\phi(z_\cdot^{[l-1]}(\vb*{x})+b_\cdot^{[l-1]})]\,,\quad \text{and}\\
k^{[l]}(\vb*{x},\vb*{x}') & = {\sigma_w^2}^{[l]}\mathbb{E}_{z_\cdot^{[l-1]},b_\cdot^{[l-1]}}[\phi(z_\cdot^{[l-1]}(\vb*{x})+b_\cdot^{[l-1]})\cdot\phi(z_\cdot^{[l-1]}(\vb*{x}')+b_\cdot^{[l-1]})]\,.
\end{split}
\end{equation}
Therefore, the neural network outputs, collected in the $(L+1)$-th layer, also follow independent, identical Gaussian process priors, written as $y_\cdot\sim \mathcal{GP}(h_\text{NN},k_\text{NN})=\mathcal{GP}(h^{[L+1]},k^{[L+1]})$. Often referred to as \emph{neural-network-induced} Gaussian processes, such priors can be explicitly formulated through the recursive relation \eqref{recursive}. 

\section{Regression using neural-network-induced Gaussian processes}

\vspace{1.5mm}
\noindent $\bullet$~~\textbf{Gaussian process regression}
\vspace{1.5mm}

\noindent As discussed in Section \ref{NNGP}, each output $y_\cdot(\vb*{x})$ follows a prior of neural-network-induced Gaussian process. Here we assume that the output is corrupted by an independent Gaussian noise, written as
\begin{equation}
y_\cdot(\vb*{x})\sim \mathcal{GP}(h_\text{NN}(\vb*{x}),k_\text{NN}(\vb*{x},\vb*{x}'))+\mathcal{N}(0,\sigma_\epsilon^2)\,.
\end{equation}
Conditioned on the training data $(\vb{X},\vb{y})=\{(\vb*{x}^{(m)},y_\cdot^{(m)})\}_{m=1}^{M}\in \mathbb{R}^{d_\text{in}\times M}\times \mathbb{R}^{M}$, the noise-free posterior output $y_\cdot^*(\vb*{x})|\vb{X},\vb{y}$ follows a new Gaussian process \cite{williams2006gaussian}, i.e., $y_\cdot^*(\vb*{x})|\vb{X},\vb{y} \sim \mathcal{GP}(h_\text{NN}^*(\vb*{x}),k_\text{NN}^*(\vb*{x},\vb*{x}'))$ whose mean and covariance functions are given as
\begin{equation}
\begin{split}
h_\text{NN}^*(\vb*{x}) & =h_\text{NN}(\vb*{x})+k_\text{NN}(\vb{X},\vb*{x})^\text{T}[\vb{K}+\sigma_\epsilon^2\vb{I}_M]^{-1}(\vb{y}-h_\text{NN}(\vb{X}))\,,\quad \text{and}\\
k_\text{NN}^*(\vb*{x},\vb*{x}') & = k_\text{NN}(\vb*{x},\vb*{x}') -  k_\text{NN}(\vb{X},\vb*{x})^\text{T}[\vb{K}+\sigma_\epsilon^2\vb{I}_M]^{-1}k_\text{NN}(\vb{X},\vb*{x}')\,,
\end{split}
\end{equation}
in which $h_\text{NN}(\vb{X}):=\{h_\text{NN}(\vb*{x}^{(1)}),\cdots, h_\text{NN}(\vb*{x}^{(M)})\}^\text{T}\in\mathbb{R}^M$, $k_\text{NN}(\vb{X},\vb*{x})=: \{k_\text{NN}(\vb*{x}^{(1)},\vb*{x}),\cdots,k_\text{NN}(\vb*{x}^{(M)},\vb*{x})\}^\text{T}\in\mathbb{R}^M$, and $\vb{K}:=k_\text{NN}(\vb{X},\vb{X})=[k_\text{NN}(\vb*{x}^{(i)},\vb*{x}^{(j)})]_{i,j=1}^{M}\in \mathbb{R}^{M\times M}$. In addition, the hyperparameters, including the noise $\sigma^2_\epsilon$ and the kernel parameters $\vb*{\theta}=\{\mu_w^{[1]},{\sigma_w^2}^{[1]},\cdots,\mu_w^{[L+1]},{\sigma_w^2}^{[L+1]};\mu_b^{[1]},{\sigma_b^2}^{[1]},\cdots,\mu_b^{[L]},{\sigma_b^2}^{[L]}\}^\text{T}$, can be determined by maximizing the log marginal likelihood as follows
\begin{equation}
\begin{split}
&\qquad \qquad \qquad \qquad \qquad \qquad \qquad (\vb*{\theta},\sigma_\epsilon^2)  = \arg\max_{\vb*{\theta},\sigma_\epsilon^2}~ \log p(\vb{y}|\vb{X},\vb*{\theta},\sigma_\epsilon^2) \\
=  &~ \arg\max_{\vb*{\theta},\sigma_\epsilon^2}\left\{ -\frac{1}{2}(\vb{y}-h_\text{NN}(\vb{X};\vb*{\theta}))^\text{T}(\vb{K}(\vb*{\theta})+\sigma_\epsilon^2\vb{I})^{-1}(\vb{y}-h_\text{NN}(\vb{X};\vb*{\theta}))-\frac{1}{2}\log |\vb{K}(\vb*{\theta})+\sigma_\epsilon^2\vb{I}| -\frac{M}{2}\log(2\pi)  \right\}\,.
\end{split}
\end{equation}

\vspace{1.5mm}
\noindent $\bullet$~~\textbf{Viewpoint of kernel ridge regression}
\vspace{1.5mm}

\noindent A reproducing kernel Hilbert space $\mathcal{H}_{k_\text{NN}}(\Omega)$ is induced by the kernel function $k_\text{NN}$ and provides a completion of the following function space of reproducing kernel map reconstruction \cite{williams2006gaussian}:
\begin{equation}
\mathcal{C}_{k_\text{NN}}=\left\{f(\vb*{x})=\sum_{m=1}^{M}\beta_m k_\text{NN}(\vb*{x}^{(m)},\vb*{x}) \Big\vert ~ M\in \mathbb{N}^+, \vb*{x}^{(m)}\in\Omega, \beta_m \in \mathbb{R}\right\}\,.
\end{equation}
Note that $\mathcal{C}_{k_\text{NN}}$ is equipped with an inner product $\langle \cdot, \cdot \rangle_{\mathcal{H}_{k_\text{NN}}} $ defined as follows: if $f(\vb*{x})=\sum_{m=1}^{M}\beta_m k_\text{NN}(\vb*{x}^{(m)},\vb*{x})$, and $g(\vb*{x})= \sum_{m'=1}^{M'}\beta'_{m'} k_\text{NN}(\vb*{x}'^{(m')},\vb*{x})$, then
\begin{equation}
\langle f, g \rangle_{\mathcal{H}_{k_\text{NN}}} = \sum_{m=1}^{M}\sum_{m'=1}^{M'}\beta_m \beta'_{m'} k_\text{NN}(\vb*{x}^{(m)},\vb*{x}'^{(m')})\,.
\end{equation}

Comparing the posterior mean $h^*_\text{NN}(\vb*{x})$ with the prior mean $h_\text{NN}(\vb*{x})$, the correction term, $\Delta_\text{NN}(\vb*{x})=h^*_\text{NN}(\vb*{x})-h_\text{NN}(\vb*{x})= k_\text{NN}(\vb{X},\vb*{x})^\text{T} \vb*{\beta} =\sum_{m=1}^{M}\beta_m k_\text{NN}(\vb*{x}^{(m)},\vb*{x})$, is evidently a reproducing kernel map reconstruction, and thus $\Delta_\text{NN} \in \mathcal{H}_{k_\text{NN}}$. Here
\begin{equation}
\vb*{\beta} = [\vb{K}+\sigma_\epsilon^2\vb{I}_M]^{-1}(\vb{y}-h_\text{NN}(\vb{X}))
= \arg\min_{\vb*{\beta}\in\mathbb{R}^M}\left\{ \|\vb{y}-h_\text{NN}(\vb{X}) -\Delta_\text{NN}(\vb{X})  \|_2^2 + \sigma_\epsilon^2 \|\Delta_\text{NN}\|_{\mathcal{H}_{k_\text{NN}}} ^2 \right\}\,,
\end{equation}
in which $\|\Delta_\text{NN}\|_{\mathcal{H}_{k_\text{NN}}} ^2 = \vb*{\beta}^\text{T}\vb{K}\vb*{\beta}$, i.e., the combination coefficients of the kernel map reconstruction are determined through a least squares problem regularized by $\|\cdot\|^2_{_{k_\text{NN}}}$, often referred to as a kernel ridge regression \cite{bartlett2021deep}. 

\section{Two-layer neural network}

In this section we consider the case when $L=1$. For the sake of conciseness, the notation of first-layer weights and biases is simplified as $\vb*{w}:=\vb*{w}_\cdot^{[1]}$ and $b:=b_\cdot^{[1]}$, respectively, and we take $\mu_{w}^{[2]}=0$ and ${\sigma_{w}^2}^{[2]}=1$ without loss of generality. 

\vspace{1.5mm}
\noindent $\bullet$~~\textbf{Reproducing kernel Hilbert space}
\vspace{1.5mm}

\noindent In such a case we have $h_\text{NN}(\vb*{x})=0$ and 
\begin{equation}
k_\text{NN}(\vb*{x},\vb*{x}')=\mathbb{E}_{(\vb*{w},b)\sim \pi}[\phi(\vb*{w}^\text{T}\vb*{x}+b)\cdot \phi(\vb*{w}^\text{T}\vb*{x}'+b)]:=k_\pi(\vb*{x},\vb*{x}')\,,
\end{equation}
in which $k_\pi$ denotes the kernel function induced by a two-layer neural network when $(\vb*{w},b)\sim \pi$. The corresponding reproducing kernel Hilbert space by $k_\pi$ is hence denoted by $\mathcal{H}_{k_\pi}(\Omega)$.

We define for a fixed $\pi\in P(\mathbb{S}^{d_\text{in}})$ that 
\begin{equation}
\begin{split}
& \mathcal{H}_\pi(\Omega) :=\left\{f(\vb*{x}) = \int_{\mathbb{S}^{d_\text{in}}} \alpha(\vb*{w},b) \phi(\vb*{w}^\text{T}\vb*{x}+b) ~\dd\pi(\vb*{w},b) \Big\vert \|f\|_{\mathcal{H}_\pi} < \infty \right\} \,,\\
& \text{with}~~ \|f\|_{\mathcal{H}_\pi}^2 := \mathbb{E}_{(\vb*{w},b)\sim \pi}[|\alpha(\vb*{w},b)|^2]\,,
\end{split}
\end{equation}
where $\mathbb{S}^{d_\text{in}}:= \{(\vb*{w},b)|~\|\{\vb*{w}^\text{T},b\}^\text{T}\|_1 = 1\}$, and $P(\mathbb{S}^{d_\text{in}})$ denotes the collection of all probability measures on $(\mathbb{S}^{d_\text{in}},\mathcal{F})$, $\mathcal{F}$ being the Borel $\sigma$-algebra on $\mathbb{S}^{d_\text{in}}$. It has been shown that $\mathcal{H}_\pi = \mathcal{H}_{k_\pi}$, and \cite{ma2018priori,rahimi2008uniform} are referred to for more details.

\vspace{1.5mm}
\noindent $\bullet$~~\textbf{Barron space}
\vspace{1.5mm}

\noindent Naturally connected with the aforementioned reproducing kernel Hilbert spaces is the Barron space \cite{ma2018priori,weinan2021barron} defined as
\begin{equation}
\begin{split}
& \mathcal{B}_2(\Omega) :=\left\{f(\vb*{x}) = \int_{\mathbb{S}^{d_\text{in}}} \alpha(\vb*{w},b) \phi(\vb*{w}^\text{T}\vb*{x}+b) ~\dd\pi(\vb*{w},b) \Big\vert \pi\in P(\mathbb{S}^{d_\text{in}}), \|f\|_{\mathcal{B}_2 } < \infty  \right\} \,,\\
& \text{with}~~  \|f\|_{\mathcal{B}_2}^2 := \inf_{\pi} \mathbb{E}_{(\vb*{w},b)\sim \pi}[|\alpha(\vb*{w},b)|^2]\,.
 \end{split}
\end{equation}
Thus we have
\begin{equation}\label{union}
\mathcal{B}_2 (\Omega)= \bigcup_{\pi\in P(\mathbb{S}^{d_\text{in}})} \mathcal{H}_\pi(\Omega) = \bigcup_{\pi\in P(\mathbb{S}^{d_\text{in}})} \mathcal{H}_{k_\pi}(\Omega)\,,
\end{equation}
i.e., the Barron space $\mathcal{B}_2$ is the union of a class of reproducing kernel Hilbert spaces $\mathcal{H}_{k_\pi}$ that are defined by the neural-network-induced kernels $k_\pi$ through two-layer neural networks. 

Upon $(\vb*{w},b)\sim \pi\in \mathbb{S}^{d_\text{in}}$, we consider an additional random variable $a$ such that $(a,\vb*{w},b)\sim \rho$, $\rho\in P(\mathbb{R}\times \mathbb{S}^{d_\text{in}})$, and $\mathbb{E}_{(a,\vb*{w},b)\sim \rho}[a|\vb*{w},b]$ $= \alpha(\vb*{w},b)$. Hence a function $f\in \mathcal{B}_2(\Omega)$ admits the form
\begin{equation}
f(\vb*{x}) = \int_{\mathbb{S}^{d_\text{in}}} \mathbb{E}_{\rho}[a|\vb*{w},b] \cdot \phi(\vb*{w}^\text{T}\vb*{x}+b) ~\dd\pi(\vb*{w},b) = \int_{\mathbb{R}\times \mathbb{S}^{d_\text{in}}} a \phi(\vb*{w}^\text{T}\vb*{x}+b)  ~\dd\rho(a,\vb*{w},b)\,.
\end{equation}
A Monte Carlo estimate with $N$ samples of $(a_i, \vb*{w}_i, b_i)$ drawn from $\rho$ is then given as
\begin{equation}\label{MC}
f(\vb*{x}) \approx \frac{1}{N} \sum_{i=1}^{N} a_i \phi(\vb*{w}_i^\text{T}\vb*{x}+b_i)\,,
\end{equation} 
which coincides with the expression of a two-layer neural network's output. Considering that the weights of first-layer neurons are $(a_i/N)$'s whose mean and variance are both proportional to $1/N$, the distribution from which $(a_i/N,\vb*{w}_i,b_i)$'s are drawn \textit{almost} aligns with the assumptions made in the beginning of Section \ref{NNGP}, except that $a_i$'s are not independent of $(\vb*{w}_i,b_i)$'s. In fact such a dependency does not impact the outputs' being Gaussian processes, as it will only impact the second layer's infinite sum which does not exist in a two-layer network. 

 As $N=N_1\to \infty$, the Monte Carlo \eqref{MC} can approximate all the functions in $\mathcal{B}_2(\Omega)$ as $\rho$ ranging over all possible options, whereas the assumption of infinite width in Section \ref{NNGP} guarantees the applicability of the central limit theorem. The kernel map constructions in all $\mathcal{H}_{k_\pi}$'s that are induced by the network, as given in \eqref{union}, and the Monte Carlo estimates sampled from all $\rho$'s, as given in \eqref{MC}, will eventually recover the same function space $\mathcal{B}_2(\Omega)$.

\bibliographystyle{abbrv}
\bibliography{refs.bib}

\begin{thebibliography}{1}

\bibitem{bartlett2021deep}
P.~L. Bartlett, A.~Montanari, and A.~Rakhlin.
\newblock Deep learning: a statistical viewpoint.
\newblock arXiv: 2103.09177, 2021.

\bibitem{durrett2019probability}
R.~Durrett.
\newblock {\em Probability: Theory and Examples}.
\newblock Cambridge university press, 2019.

\bibitem{ma2018priori}
W.~E, C.~Ma, and L.~Wu.
\newblock A priori estimates of the population risk for two-layer neural
  networks.
\newblock arXiv: 1810.06397, 2018.

\bibitem{weinan2021barron}
W.~E, C.~Ma, and L.~Wu.
\newblock The $\text{B}$arron space and the flow-induced function spaces for
  neural network models.
\newblock {\em Constructive Approximation}, pages 1--38, 2021.

\bibitem{lee2017deep}
J.~Lee, Y.~Bahri, R.~Novak, S.~S. Schoenholz, J.~Pennington, and
  J.~Sohl-Dickstein.
\newblock Deep neural networks as $\text{G}$aussian processes.
\newblock arXiv: 1711.00165, 2017.

\bibitem{neal2012bayesian}
R.~M. Neal.
\newblock {\em Bayesian Learning for Neural Networks}.
\newblock Springer Science \& Business Media, 2012.

\bibitem{rahimi2008uniform}
A.~Rahimi and B.~Recht.
\newblock Uniform approximation of functions with random bases.
\newblock In {\em 2008 46th Annual Allerton Conference on Communication,
  Control, and Computing}, pages 555--561. IEEE, 2008.

\bibitem{strang2019linear}
G.~Strang.
\newblock {\em Linear Algebra and Learning from Data}.
\newblock Wellesley-Cambridge Press, 2019.

\bibitem{williams2006gaussian}
C.~K. Williams and C.~E. Rasmussen.
\newblock {\em Gaussian Processes for Machine Learning}.
\newblock MIT press Cambridge, MA, 2006.

\end{thebibliography}

 \vspace{5mm}

\noindent \textit{A preliminary version of this note was used for the internal discussions within the Dynamics \& Deep Learning Reading Club at the University of Twente.}

\end{document}